\PassOptionsToPackage{table}{xcolor}

\documentclass[sigconf,natbib=true]{acmart}

\setcitestyle{square,sort,comma,numbers}

\usepackage{times}
\usepackage{tcolorbox}
\usepackage{pifont}
\usepackage[utf8]{inputenc} 
\usepackage[T1]{fontenc}    
\usepackage{hyperref}
\usepackage{url}            
\usepackage{booktabs}       
\usepackage{amsfonts}       
\usepackage[show]{notes-alt}
\usepackage{nicefrac}       
\usepackage{microtype}      
\usepackage{xcolor} 
\usepackage[table]{xcolor}
\definecolor{darkblue}{rgb}{0.0, 0.0, 0.55}
\definecolor{lightblue}{rgb}{0.68, 0.85, 0.9}
\definecolor{lightcyan}{rgb}{0.88, 1.0, 1.0}
\definecolor{darkcerulean}{rgb}{0.03, 0.27, 0.49}
\usepackage{amstext}
\usepackage{amsmath}
\usepackage{enumitem}
\usepackage{algorithm}
\usepackage{algpseudocode}
\usepackage{dblfloatfix}
\usepackage{multirow}
\usepackage{wrapfig}
\usepackage{tikz}
\usepackage{pgfplots}
\usepackage{subcaption}
\usepackage{pgfplotstable}
\usepackage{pgfplots}
\pgfplotsset{compat=1.15}
\usepgfplotslibrary{
  statistics,
  colorbrewer,
  groupplots,
}
\usetikzlibrary{
  patterns,
  shapes.geometric,
  decorations.text,
  matrix,
  fit,
  backgrounds,
  positioning,
}
\tikzset{
  fignode/.style={
    outer sep=0.25em,
  }
}
\tikzset{
  framedfignode/.style={
    outer sep=0.25em,
    inner sep=0.5em,
    rounded corners,
    draw,
  }
}
\colorlet{plotColorNeutral}{gray}
\definecolor{plotColor1}{HTML}{f61a1c}
\definecolor{plotColor2}{HTML}{377eb8}
\definecolor{plotColor3}{HTML}{4daf4a}
\definecolor{plotColor4}{HTML}{984ea3}
\colorlet{plotColorNeutral*}{plotColorNeutral!40}
\colorlet{plotColor1*}{plotColor1!60}
\colorlet{plotColor2*}{plotColor2!60}
\colorlet{plotColor3*}{plotColor3!60}
\colorlet{plotColor4*}{plotColor4!60}
\pgfplotsset{
    colormap={greenred}{HTML=(4daf4a) HTML=(e41a1c)},
    colormap={redgreen}{HTML=(e41a1c) HTML=(4daf4a)}
}

\newcommand{\knn}{\textsc{knn}}

\newcommand{\COT}{\textsc{Few-shot-cot}}

\newcommand{\fshot}{\textsc{few-shot}}

\newcommand{\sys}[1]{\textsc{Gar}\def\temp{#1}\ifx\temp\empty{}\else\raisebox{-.4ex}{\scriptsize #1}\fi}

\newcommand{\sgar}[1]{\textsc{Sgar}\def\temp{#1}\ifx\temp\empty{}\else\raisebox{-.4ex}{\scriptsize#1}\fi}

\newcommand{\sysbm}[1]{\sys{}${}_{BM25}$}
\newcommand{\quamsys}[1]{\textsc{Quam}\def\temp{#1}\ifx\temp\empty{}\else\raisebox{-.4ex}{\scriptsize #1}\fi}

\newcommand{\cerberussys}[1]{\textsc{Ore}\def\temp{#1}\ifx\temp\empty{}\else\raisebox{-.4ex}{\scriptsize #1}\fi}
\newcommand{\cerberuhalfssys}[1]{\textsc{CerberusHalf}\def\temp{#1}\ifx\temp\empty{}\else\raisebox{-.4ex}{\scriptsize #1}\fi}

\newcommand{\name}{\textsc{TLQA}}

\newcommand{\argmaxm}[1]{%
  \ifthenelse{\isempty{#1}}%
    {\overset{m}{\argmax}}
    {\underset{#1}{\overset{m}{\argmax}}\, }
}
\newcommand{\argminm}[1]{%
  \ifthenelse{\isempty{#1}}%
    {\overset{m}{\argmin}}
    {\underset{#1}{\overset{m}{\argmin}}\, }
}

\usepackage{pifont}

\copyrightyear{2025}
\acmYear{2025}
\acmConference[ICTIR '25]{Proceedings of the 2025 International ACM SIGIR
Conference on Innovative Concepts and Theories in Information Retrieval
(ICTIR)}{July 18, 2025}{Padua, Italy}
\acmBooktitle{Proceedings of the 2025 International ACM SIGIR Conference on
Innovative Concepts and Theories in Information Retrieval (ICTIR) (ICTIR '25), July
18, 2025, Padua, Italy}
\acmISBN{979-8-4007-1861-8/2025/07}

\begin{document}

\title{Evaluating List Construction and Temporal Understanding capabilities of Large Language Models}

 \author{Alexandru Dumitru$^*$}
  \email{alexandru.dumitru@prosus.com}
 \affiliation{
   \institution{Prosus}
   \city{Delft}
   \country{Netherlands}
 }
 
  \author{Venktesh V$^*$}
  \email{v.viswanathan-1@tudelft.nl}
 \affiliation{
   \institution{Delft University of Technology}
   \city{Delft}
   \country{Netherlands}
 }
 
 \author{Adam Jatowt}
  \email{adam.jatowt@uibk.ac.at}
 \affiliation{
   \institution{University of Innsbruck}
   \city{Innsbruck}
   \country{Austria}
 }

 \author{Avishek Anand}
  \email{avishek.anand@tudelft.nl}
 \affiliation{
   \institution{Delft University of Technology}
   \city{Delft}
   \country{Netherlands}
 }
\begin{abstract}

\end{abstract}


\begin{abstract}

Large Language Models (LLMs) have demonstrated immense advances in a wide range of natural language tasks. However, these models are susceptible to hallucinations and errors on particularly temporal understanding tasks involving multiple entities in answers. In such tasks, they fail to associate entities with accurate time intervals, generate a complete list of entities in answers or reason about events associated with specific temporal bounds. Existing works do not extensively evaluate the abilities of the model to perform implicit and explicit temporal understanding in a list answer construction setup.  To bridge this gap, we propose the \textit{\textbf{T}ime referenced \textbf{L}ist based \textbf{Q}uestion \textbf{A}nswering } or \name{} benchmark that requires structured answers in list format aligned with corresponding time periods. 
Our \name{} benchmark, requires both list construction and temporal understanding simultaneously, which to the best of our knowledge has not been explored in prior benchmarks. We investigate the temporal understanding and list construction capabilities of state-of-the-art generative models on \name{} in closed-book and open-domain settings. 
Our findings reveal significant shortcomings in current models, particularly their inability to provide complete answers and temporally align facts in a closed-book setup and the need to improve retrieval in open-domain setup, providing clear future directions for research on \name{}. The benchmark and code at \url{https://github.com/elixir-research-group/TLQA}.

\end{abstract}

\maketitle

\def\thefootnote{*}\footnotetext{These authors contributed equally to this work}\def\thefootnote{\arabic{footnote}}
\section{Introduction}
\label{chapter:introduction}


Large Language Models have made tremendous advances in wide range of Natural language processing (NLP) tasks \cite{brown2020gpt3,gpt3incontext,openai2023gpt4}. Their ability to reason and answer questions with factual information has been studied widely through well known Question Answering (QA) benchmarks \cite{baseball, rajpurkar-etal-2016-squad,liu-etal-2018-stochastic,DBLP:journals/corr/abs-2305-09617/medpalm2}. However, such models are still prone to errors and hallucination \cite{hallucination} where they generate plausible sounding answers not grounded on facts in the real world. 
\begin{table}[htb!!]
\begin{tcolorbox}[title=Example: \name{} and extensions, colframe=darkcerulean,
colback=lightcyan
]
\small
\textbf{\name{}}: \textit{List all political positions Joe Biden held to this day.}
\\
\textbf{[Answer]:} \\
$\text{\rlap{$\checkmark$}}\square$
 President of the United States (2021-2024) \newline
$\text{\rlap{$\checkmark$}}\square$
 Vice President of the United States (2009-2017) \newline
$\text{\rlap{$\checkmark$}}\square$
 United States Senator from Delaware (1973-2009) \\
 \textbf{\name{}-TS question}: \textit{What political positions were held by Biden between 2009-2020?} \\
 \textbf{[Answer]:} $\text{\rlap{$\checkmark$}}\square$
 Vice President of the United States (2009-2017) 

  \textbf{\name{}-TM question}: \textit{What political positions were held by Biden after serving as Senator ?} \\
 \textbf{[Answer]:} $\text{\rlap{$\checkmark$}}\square$
 Vice President of the United States (2009-2017)
$\text{\rlap{$\checkmark$}}\square$
 President of the United States (2021-2024) \newline

\end{tcolorbox}
\captionof{figure}{An example for Time referenced List based QA}
\label{fig:examples}
\end{table}
These problems are further exacerbated for questions where the model has to generate a structured list of answers (\textbf{ListQA}). In such ListQA tasks, the model has to ensure coverage of all relevant entities in the answer (list construction). The ListQA task may include inquiries referring to side effects of certain medication, historical events, priority lists  or rankings. 
Over \textbf{10\% of Bing's web queries} \cite{DBLP:journals/corr/abs-2001-04828/tableQA} and healthcare queries \cite{DBLP:journals/bioinformatics/YoonJLK22/bioeqa}  \textbf{21.9\%} of the queries necessitate a structured list as response. However, there have been only few works on queries that require structured, listwise answers \cite{mavi2022survey}.

Additionally, factual information of entities evolves over time and requires reasoning about the scope of knowledge in different time periods. Hence, the LLM must posses \textbf{temporal understanding} capabilities such as associating the entities in the list answer with accurate time intervals and also reason about temporal scope of the entity in the question. This may also entail performing implicit temporal understanding for questions like ``What positions were held by Joe Biden after his tenure as senator?". Here the LLM has to infer the time period of position as senator and perform temporal arithmetic to arrive at the time period after the official left this position to arrive at the correct answer. While existing works have focused on questions with temporal markers, they were primarily based on Knowledge Graphs (KG) \cite{tempwebquestions, Jia_2021, chen-etal-2023-multi,mavromatis2021tempoqr,shang-etal-2022-improving} and not text-based, which limits the understanding of temporal evolution and transitions in text. The emphasis on KG based temporal facts has been found to be limiting research as it ignores implicit temporal structure or temporal reasoning tasks beyond KGs. Further, prior works have observed that \textbf{heuristics or shortcuts} can often answer these questions without necessitating genuine temporal reasoning \cite{chen_subgraph}.  While few works focus on temporal understanding over natural text \cite{dhingra2022time/templama,margatina2023dynamictemplama}, these works do not address questions that require multiple answers (list based), which is common in real-world scenarios involving time-period-specific queries.




To bridge this gap, we study time-referenced List based QA, by curating a benchmark (\name{}) along with \textbf{large evidence collection} of Wikipedia articles and corresponding info boxes. The questions in \name{} require a list of answers each associated with a time period. This reflects real-world queries where users are interested in querying historical events or news \cite{wang2022archivalqa}. These are common queries issued to search engines  which are time-sensitive or involve temporal specifiers \cite{chen2021dataset, tempwebquestions} such as ``What teams did Lebron James play for between 2007-2009?". These queries are complex to resolve as they need to not only maintain completeness of the answer list (\textbf{list construction} ability) but also get accurate bounds (start and end) of the related time periods and ability to associate entities with the accurate time periods (\textbf{temporal understanding} ability). Some queries in our benchmark, such as the \name{}-TM question in Figure 1 also require implicit temporal understanding capability to infer time period related to the event specified in the question. Examples of diverse question types in our benchmark are presented in Figure 1. Unlike typical QA scenarios such as those in SQuAD \cite{DBLP:journals/corr/RajpurkarZLL16/squad} where the system extracts a single answer from a single body of text, \name{} systems must find multiple answers from a single text or multiple sources \cite{DBLP:journals/corr/abs-2306-00435/HowMany} and align them with appropriate time periods while ensuring the bounds for time intervals are accurate. 
We would also like to note that exact sorting of time periods is beyond scope, though our benchmark can be used for the same.
We evaluate a range of generative Large Language Models (LLMs) in few-shot closed book setup and in open-domain setup to analyze their \textbf{temporal understanding} and their \textbf{list construction} abilities.  
To the best of our knowledge, prior works have not tackled the temporal understanding and list structured output abilities simultaneously, critical for real-world applications.

\textbf{Real-world applications:} Benchmarking LLMs on TLQA can help improve their temporal understanding and list construction capabilities, which is of immense use in healthcare, historical research and education and journalism. For instance, in historical research/education it helps scholars access precise historical information regarding different events across time. In healthcare, it could be used to access historical medical records. 

\textbf{Research Questions:}


 \textbf{RQ1:} How well do Large Language Models (LLMs) perform on temporal understanding and list construction based questions in \name{} ?
 
 \textbf{RQ2}: Does retrieval augmentation from external knowledge sources help reduce hallucination in LLMs for generating complete lists in answers with precise temporal bounds?
 
 \textbf{RQ3}: When provided with golden evidence in retrieval augmented setup, what are the effects of various distractors (as evidence) on model performance?






\vspace{-0.5em}
\section{Related Work}
\label{sec:rel-work}

While several ListQA benchmarks have been proposed, they primarily focus on answering factoid or ambiguous questions with multiple answers \cite{DBLP:journals/corr/abs-2205-12665/qampari,DBLP:conf/aaai/LeeKK23Liquid,DBLP:journals/corr/abs-2210-14353/romqa}. They do not focus on the evolution of such factual answers with shift in temporal information.  
Temporal information plays an important role for real-world tasks such as Information Retrieval and Question Answering\cite{chen2021dataset, zhang-etal-2024-analyzing}. Information evolves over time \cite{10.1145/1963192.1963296, gottschalk2018eventkg}, and it is critical to provide temporally recent and relevant information to users. Hence, temporal information retrieval \cite{temporal_ir_1,temporal_ir_2} and temporal QA \cite{shang-etal-2022-improving,son-oh-2023-time} has been of immense interest recently. These approaches commonly use temporal signals in text to ascertain temporal aspects of query intents \cite{temporal_intent}, perform query or document matching \cite{time_aware_document_retrieval}, aiding in search of web archives\cite{temporal_web_search}.  A large body of studies have been dedicated to study evolution of facts through temporal knowledge graphs (TKG) \cite{Jia_2021,tempwebquestions,zhang-etal-2024-mustq,10135342} and evaluate QA over such graphs \cite{saxena-etal-2021-question}. However, \cite{chen_subgraph} observed that such datasets consists of primarily pseudo-temporal questions where QA tasks could be solved without enforcing temporal constraints. To tackle this, MultiTQ \cite{chen-etal-2023-multi} mandates temporal constraints for QA.

 However, answering questions over temporal knowledge graphs are limited to the facts contained in the constructed knowledge graph. To study temporal evolution of facts in natural text, several datasets \cite{chen2021dataset, Kasai2022RealTimeQW} focus on time aware QA with textual evidence.
 \begin{figure*}
 \centering
\includegraphics[width=0.8\textwidth]{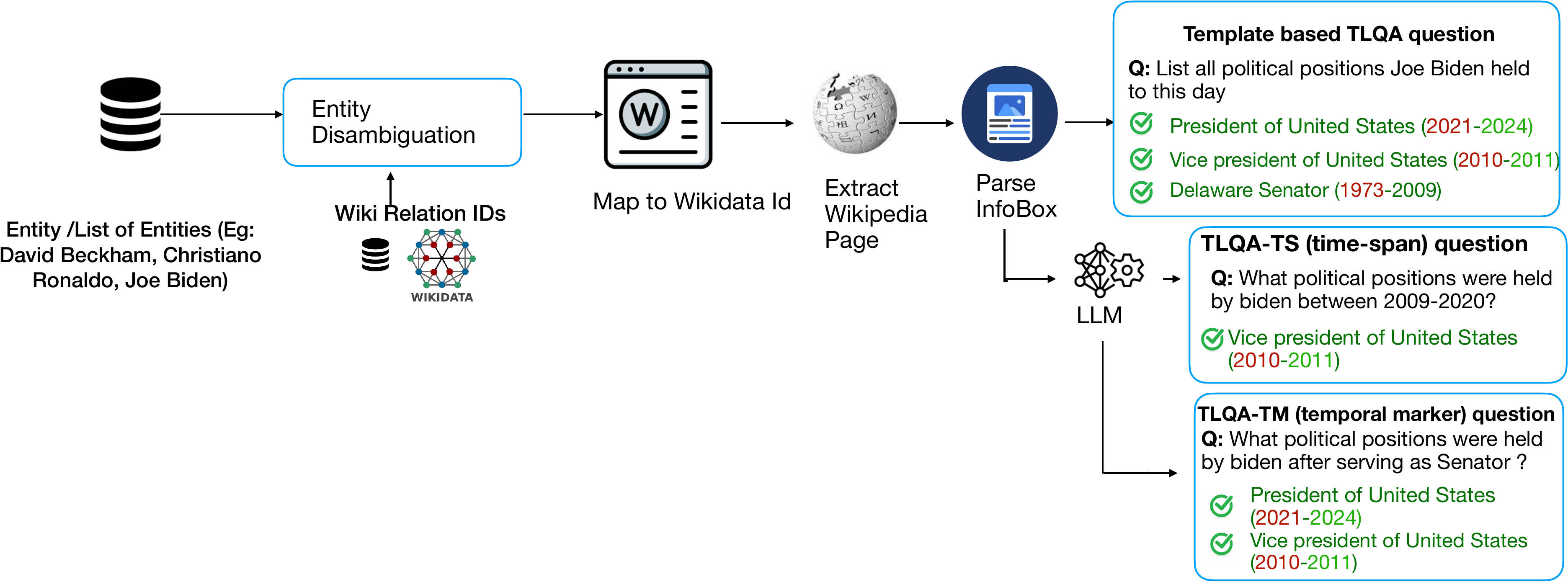}    
\caption{\name{} data collection pipeline}
\label{fig:pipeline}
\end{figure*}
 More recently, \textbf{TempLAMA} \cite{dhingra2022time/templama} introduced cloze type questions to test the temporal awareness of pre-trained language models. However, it has several limitations, such as answers with missing start date and  7.9\% of the questions have subjects with incomplete names, which can
be problematic for disambiguation such as ’Cristiano Ronaldo’ and ’Ronaldo,’. Additionally, the evaluation metrics employed in TempLAMA such as maximum token-level F1 score does not account for completeness of answers and the correctness of temporal bounds. 
 

\vspace{-0.7em}
\section{Benchmark Creation}

\subsection{Time Referenced List based QA}

 \begin{definition}
A time-referenced list based question $Q$ is a query that, given  a temporal context $t$, requests a comprehensive list of entities or facts $e$ constituting together a correct answer to $Q$ over the time period $t$. The answer $A$ to $Q$ is a set of pairs $(e,\tau)$ where $e$ is an entity or fact, and $\tau$ is the time interval when $e$ is relevant to $Q$.
\end{definition}


\subsection{TLQA generation}

We propose an automated solution to generate \name{}. Our focus is on generating questions from entities related to Wikidata relations namely P54 (member of a sports team) and P39 (position held). The main intuition behind this decision is that only these topics, the entities are naturally associated with multiple organizations/positions over different time periods, making it a natural and only choice for \name{}. Our data-curation pipeline consists of multiple stages such as entity/subject extraction, Wikipedia Mapping, Infobox Extraction and question generation using templates as shown in Figure \ref{fig:pipeline}.

\subsubsection{Entity seed set}
Our pipeline first starts with a list of entities/subjects which would form the core of the questions. We extract these entities from TempLAMA to act as \textbf{seed set} though the pipeline would work with any set of entities. Hence, we would like to note that our pipeline is generic \textbf{without dependency} on TempLAMA and requires only a list of entities or subjects with relation pairs. For each query in TempLAMA, the subject of each query is extracted by using the Wikidata relation type as a semantic marker. For example, for relation P54, each query is of the form  \texttt{subject plays for \_X\_}, and can be split based on the words \texttt{plays for}, to extract the subject. 
\subsubsection{Entity Disambiguation and mapping to Wikidata ID}
For each subject/entity, the goal is to map to Wikipedia page to extract relevant information and generate questions. Using only the subject's name may be insufficient, as multiple named entities may share the same name. Therefore, to resolve each subject to its correct Wikipedia page, we collect Wikidata ID which provides a  one-to-one relationship between entities and Wikipedia articles.

To accurately find the Wikidata ID for each subject, we utilize the entity, the Wikidata relation type (e.g., member of sports teams), and the aggregated set of entities related to the subject retrieved using Wikidata relation type, with each related entity having a unique Wikidata identifier. For example, related entities in the context of a sports celebrity could be some of the teams he played for which are objects of the relation type ``member of sports teams" . We then search for entities with the subject's name on Wikidata, limiting the search to the top 50 results to manage the scope. Next, we filter these results to retain only those entities that have the specified relation type (e.g., P54 for sports teams). Based on the overlap between object IDs in relations for the above candidates and IDs of related entities extracted earlier for the subject, the Wikidata ID is assigned. 

\begin{figure}[!t]
    \begin{subfigure}{0.58\linewidth}
    \begin{tikzpicture}

    \begin{axis}[
            ybar=2pt,
            width=0.9\textwidth,
            height=4.2cm,
            bar width=0.25,
            xticklabel style={rotate=45},
            every axis plot/.append style={fill},
            grid=major,
            xtick={1, 2, 3 , 4 ,5, 6 ,7},
            xticklabels={Official, Religion, Bishop, Military, Cardinal, Scientist, Other},
            xlabel={Infobox type},
            ylabel style = {font=\tiny},
            xlabel style = {font=\tiny},
        yticklabel style = {font=\boldmath \tiny,xshift=0.25ex},
        xticklabel style ={font=\tiny,yshift=0.75ex},
            ylabel={Count},
            enlarge x limits=0.08,
            ymin=0,
            ymax=1200
        ]
        \addplot+[
            ybar,
            plotColor1*,
            draw=black,
            postaction={
                    pattern=north east lines
                },
        ] plot coordinates {
                (1,1065)
                (2,58)
                (3,19)
                (4,16)
                (5,12)
                (6,10)
                (7,136)
};
    \end{axis}
\vspace{-1em}

\end{tikzpicture}
    \begin{tikzpicture}

    \begin{axis}[
            ybar=2pt,
            width=0.9\textwidth,
            height=4.2cm,
            bar width=0.25,
            xticklabel style={rotate=45},
            every axis plot/.append style={fill},
            grid=major,
            xtick={1, 2, 3 , 4 ,5, 6 ,7},
            xticklabels={Football, Cricket, Basketball, NFL, Cyclist, Baseball, Rugby, Racing, other},
            xlabel={Infobox type},
            ylabel style = {font=\tiny},
            xlabel style = {font=\tiny},
        yticklabel style = {font=\boldmath \tiny,xshift=0.25ex},
        xticklabel style ={font=\tiny,yshift=0.75ex},
            ylabel={Count},
            enlarge x limits=0.08,
            ymin=0,
            ymax=800
        ]
        \addplot+[
            ybar,
            plotColor1*,
            draw=black,
            postaction={
                    pattern=north east lines
                },
        ] plot coordinates {
                (1,684)
                (2,99)
                (3,90)
                (4,36)
                (5,23)
                (6,22)
                (7,16)
};
    \end{axis}
\vspace{-1em}

\end{tikzpicture}
        \end{subfigure}

\caption{Infobox Type Distribution for P39 and P54}
\label{fig:infobox_distribution}
\vspace{-2em}
\end{figure}
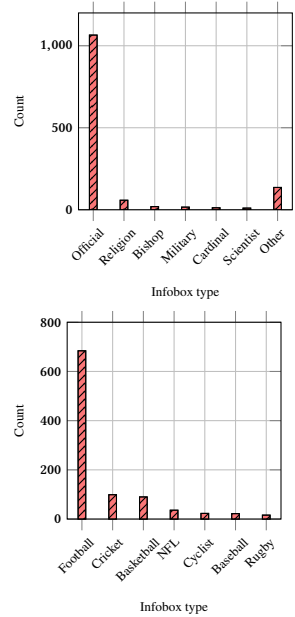
After assigning Wikidata IDs, we perform a verification step that checks the label and aliases for each assigned Wikidata ID and compare them with the normalized subject name. If a match is not found among the names, the entity is ignored. This process resulted in the reduction of the initial 5825 entities to 5711 entities.
\vspace{-0.7em}
\subsubsection{Infoboxes}
After obtaining the Wikidata IDs, we fetch the corresponding infobox for each entity from corresponding Wikipedia pages by parsing the markup. An infobox is a structured table in a Wikipedia article that presents key information about the subject in a standardized format. In Wikipedia's Wikitext, the markup language used by Wikipedia, infoboxes follow specific templates designed for different types of subjects. As of today, there are more than 1000 types of infoboxes, which can be found on Wikipedia\footnote{\url{https://en.wikipedia.org/wiki/Wikipedia:List_of_infoboxes}}. These structures often contain temporal information, which can be collected to form questions that require temporal understanding. 

\subsubsection{Infobox Distribution over Wikidata relations}
 We analyze and plot the distribution of infobox types. Infobox types that appear fewer than 10 times are grouped under 'Other'. The plots for some relations are available in Figure \ref{fig:infobox_distribution}.  We observe that, for the relation P39, \textbf{infobox officeholder (official)} represent the most common infobox type, whereas for P54, the most frequent one is \textbf{football biography (football)}. 



\subsubsection{TLQA from infoboxes}
 

\begin{table}[hbt!]
\vspace{-1em}
    \caption{Mapping from subject to query based on Wikidata relation type and infobox type}

    \centering
    \small
    \begin{tabular}{ l c c }
        \toprule
        \textbf{Infobox Type} & \textbf{Temporal Markers} & \textbf{Answer Fields} \\
        \midrule
        Football  & youth years,years & youth clubs,clubs \\ 
       Biography &national years & national team \\ 
        \midrule 
        Officeholder & term start & office,suboffice \\ 
        & term end & jr/sr,state senate \\ 
        & subterm & state assembly \\ 
        \midrule
        Cricketer & year, international span & club, country \\ 
        \bottomrule
    \end{tabular}

    \label{tab:TLQA_infobox_fields}
\end{table}
    
\noindent To form TLQA, we construct a question that explicitly requests a list of entities or facts with corresponding time periods associated with the entities. The question templates for infobox types can be seen in Table  \ref{tab:subject_to_query_mapping}.

\begin{table}[h!]
    \caption{Fields for temporal information from Infoboxes}
         \scalebox{0.8}{

    \begin{tabular}{>{\centering\arraybackslash}m{0.12\textwidth} >{\centering\arraybackslash}m{0.15\textwidth} >{\arraybackslash}m{0.27\textwidth}}
        \toprule
        \textbf{Relation ID} & \textbf{Infobox Type} & \textbf{Query} \\
        \midrule
        \multirow{2}{*}{P54} & Football,Cricketer &   \texttt{List all teams \emph{<subject>} played for to this day.} \\
        \midrule
        P39 & Officeholder &  \texttt{List all political positions \emph{<subject>} held to this day.} \\
        \bottomrule
    \end{tabular}}

    \label{tab:subject_to_query_mapping}
\end{table}

To create our answer set, given a generated query, we parse the infobox corresponding to each subject and query to extract the relevant information. We utilize the temporal markers mentioned in Table \ref{tab:TLQA_infobox_fields} to compile comprehensive answers. For sports-related queries, we extract the teams and the corresponding years the subject played for each team. For political-related queries, we extract the political positions held and the associated time periods. If the end year for a position or team association is not specified, we interpret this as an indication that the subject currently holds the position or remains with the team, aligning with the convention used on Wikipedia. 


\begin{table}[ht]
\caption{Dataset statistics along with categories.}
\small
\renewcommand{\arraystretch}{1} 
\setlength{\tabcolsep}{6pt} 
\begin{tabular}{ l l c c }
\toprule
\textbf{Category} & \textbf{Split} & \textbf{\# of entries} & \textbf{Mean \# of Answers}  \\ \midrule
\multirow{3}{*}{Political} & Training & 630 & 5.473 \\ 
 & Test & 251 & 5.438\\ 
 & Total & 881 & 5.463 \\ \midrule
\multirow{3}{*}{Sports} & Training & 528 & 11.775 \\ 
 & Test & 246 & 11.915 \\ 
 & Total & 774 & 11.820 \\ \bottomrule
\end{tabular}

\label{tab:size_mean_median_answers}
\end{table}

The process of generating questions and answers is applied to the extracted TempLAMA subjects marked with the specified infobox types. This results in a collection of \textbf{1655} questions, with an average number of answers of \textbf{8.641}. We perform a stratified train-test split based on the question's topic (either political or sports-related). The dataset statistics are shown in Table \ref{tab:size_mean_median_answers}.
\begin{table}[t!]
\caption{Manual eval. GAC: Ground Truth Answer Completeness, QU : Question usefulness. We use the Likert scale (1-5) and Cohen's Kappa ($\kappa$) for inter-annotator agreement (in brackets).}
\centering
\small
\begin{tabular}{lc}
\hline
\textbf{GAC}  ($\kappa$)& \textbf{QU} ($\kappa$) \\
\hline

\midrule
 4.75$\pm$0.50 (0.67) & 4.98$\pm$0.05 (0.74) \\

 
\bottomrule
\end{tabular}

\label{tab:question_quality}
\end{table}

\begin{table}[htb!!]
\begin{tcolorbox}[title=Prompt: Generate Temporal-Span based QA \name{}-TS, colframe=darkcerulean,
colback=lightcyan]
\small
\textbf{\ding{228} \textsc{System Message}}:
\begin{itemize}[leftmargin=1em]
    \item You will receive an original question along with its correct answer. The question asks for a list of entities associated with a person, including temporal information.
    \item Your task is to \textbf{generate a new question-answer pair} by rephrasing the original question to include a specific time interval condition, such as "between 2000 and 2010", "before 2000", or "after 2010".
    \item Select a time interval that includes as many entities as possible from the original answer (i.e., maximize overlap with the timelines). The answer should be a subset of the original answer, including only those entities whose time spans overlap with the selected interval.
    \item The new question should require temporal reasoning to answer due to the added time constraints.
    \item The answer should maintain the same format as the original answer: a list of entities with years denoting the time span.
\end{itemize}

\vspace{0.3cm}
\textbf{Output Format}: \\
Provide the response in JSON format, adhering to the schema below:
\begin{verbatim}
{ "question": "<new question>",
    "answers": { "<entity_1>": "<years>",
        "<entity_2>": "<years>",  ...}}
\end{verbatim}

\end{tcolorbox}
\captionof{figure}{Prompt used to generate a new timeline dataset.}
\label{fig:time-interval-creations-prompt}
\end{table}

\begin{table}[htb!!]
\begin{tcolorbox}[title=Prompt: Temporal Marker based Generation (\name{}-TM),colframe=darkcerulean,
colback=lightcyan]
\small
\textbf{\ding{228} \textsc{System Message}}: 
\begin{itemize}[leftmargin=1em]
    \item You will receive an original question along with its correct answer. The question asks for a list of entities associated with a person, including temporal information.
    \item Your task is to \textbf{generate a new question-answer pair} by rephrasing the original question in a more indirect way, incorporating a temporal marker based on one of the entities from the original answer.
    \item The new question should reference an event or time associated with an entity from the original answer (e.g., "after he left [Entity]", "before joining [Entity]", "following his tenure at [Entity]"). Then ask for entities that are before or after that temporal marker.
    \item The answer should be a subset of the original answer, including only those entities that satisfy the new temporal conditions specified in the question.
    \item The new question should not contain a time span, only the temporal marker.
\end{itemize}

\vspace{0.2cm}
\textbf{Output Format}: \\
Provide the response in JSON format, adhering to the schema below:
\begin{verbatim}
{"question": "<new question>",
    "answers": {
        "<entity_1>": "<years>",
        "<entity_2>": "<years>", ... }}
\end{verbatim}

\end{tcolorbox}
\captionof{figure}{Prompt to generate indirect temporal Q-A pairs.}
\label{fig:temporal_markers}
\vspace{-1em}
\end{table}

\begin{table}[htb!!]
\begin{tcolorbox}[title=Prompt: Evaluate Generated Question-Answer Pairs, colframe=darkcerulean,
colback=lightcyan]
\small
\textbf{\ding{228} \textsc{System Message}}: \\

\textbf{Objective:}
\begin{itemize}[leftmargin=1em]
    \item Determine if the generated question-answer pair is accurate, using the original pair as the baseline for truth.
    \item Pay extra attention to ensure that the years in the generated answer match the original answer.
\end{itemize}

\vspace{0.3cm}
\textbf{Response Instructions:}
\begin{itemize}[leftmargin=1.5em]
    \item Respond with:
    \begin{itemize}[leftmargin=2em]
        \item \texttt{'1'} if the new question-answer pair is correct.
        \item \texttt{'0'} if the new question-answer pair is incorrect.
    \end{itemize}
    \item Provide reasoning for your evaluation.
    \item Format the output as a JSON object.
\end{itemize}

\vspace{0.3cm}
\textbf{Output Format}: \\
Your output should be a JSON object structured as follows:
\begin{verbatim}
{"reasoning": <str, "Your reasoning for the 
answer">,"is_correct": <int, 1 if the new
    question-answer pair is correct,
    0 if it is incorrect>}
\end{verbatim}
\end{tcolorbox}
\captionof{figure}{Prompt used for ensuring the quality of automatically generated QA pairs.}
\label{fig:filter data}
\end{table}

\subsubsection{Extending TLQA}

Since the questions in the benchmark are created based on templates and centered on entities, they may have limited temporal variability and may also not test for implicit temporal understanding capabilities. To address this, we extend TLQA with two additional evaluation subsets (test sets) by generating variations of original questions in TLQA test set. These subsets namely \name{}-TS where TS denotes Time Span and \name{}-TM which denotes Temporal Markers test the ability of the model to reason and provide list-structured answers for different time slices and ability of the model to decipher implicit temporal references and perform temporal understanding respectively. We employ the powerful LLMs GPT4o and  using the prompts in Figure \ref{fig:temporal_markers} (for TM) and Figure \ref{fig:time-interval-creations-prompt} (for TS) we generate the new sets of questions. We perform data filtering using the prompt shown in Figure \ref{fig:filter data}, with a powerful GPT4o model as it shown to correlate well with human judgments \cite{verga2024replacingjudgesjuriesevaluating} to validate the correctness of the newly generated questions and answers. 

\textbf{Manual Evaluation:} We also perform manual evaluation on 100 questions sampled in a stratified manner by asking annotators to rate the question quality/usefulness and ground truth answer completeness (guidelines in repository) of the generated questions using Likert scale (1-5). The results are shown in Table \ref{tab:question_quality} along with agreement between annotators. The analysis helps validate the quality of our data curation pipeline.  This evaluation helps us further filter down to questions that are of high quality. This results in evaluation sets with \name{}-TS comprising \textbf{423} and \name{}-TM comprising \textbf{460} questions in addition to the 1655 questions mentioned earlier.

\subsection{Corpus Creation for Open-Domain Setup}
\label{sec:corpus_creation}


We release a corpus which is the latest Wikipedia dump (April 2024)\footnote{\url{https://dumps.wikimedia.org/enwiki/latest/enwiki-latest-pages-articles.xml.bz2}} to evaluate models on TLQA in an open-domain setup.  We employ dumpster-dip \cite{dumpster-dip} to parse the Wikipedia dump and extract semi-structured data such as infoboxes from the articles. After filtering out the articles without an infobox, we get a collection of approximately \textbf{4.5 million} articles. For each article, we save the title, infobox, and summary ( 'lead'\footnote{\url{https://en.wikipedia.org/wiki/Wikipedia:Manual_of_Style/Lead_section}} section of each article). 
\section{Experimental Setup}
 \label{sec:experiments}
 We evaluate several LLMs on \name{} (Mistral v0.2, Llama 3.1 8b and GPT-4o-mini). Llama 3.1 underperforms when compared to Mistral v0.2 and hence is not attached in this paper owing to space constraints. However, Llama 3.1 8b results can be found in the repo. 
 We would like to highlight that an \textit{exhaustive evaluation} of all LLMs in the current landscape is not feasible.

\begin{table*}[ht]
    \centering
    \small
    \caption{Results of Mistral v0.2 and gpt-4o-mini  models on \name{}. The first three metrics evaluate \textbf{list completeness}, the last two assess the \textbf{temporal correctness} (Temporal Overlap (TO) and Temporal Jaccard (TJ))}
    \begin{tabular}{
        >{\raggedright\arraybackslash}p{3cm}
        >{\raggedright\arraybackslash}p{2cm}
        >{\raggedright\arraybackslash}p{2cm}
        >{\raggedright\arraybackslash}p{2cm}
        >{\raggedright\arraybackslash}p{2cm}
        >{\raggedright\arraybackslash}p{2cm}
    }
        \toprule
        \textbf{Model} & \textbf{Precision} & \textbf{Recall} & \textbf{F1} & \textbf{TO} & \textbf{TJ} \\
        \midrule
        \multicolumn{6}{>{\columncolor[gray]{0.93}}l}{\textbf{Closed Book }} \\
{\textbf{Mistral v0.2 7b}} \\
         \fshot{} & 0.558 & 0.257 & 0.330 & 0.384 & 0.317 \\
         \knn{} \fshot{} & 0.532 & 0.310 & 0.366 & 0.435 & 0.360 \\
         \COT{}  & 0.529 & 0.276 & 0.338 & 0.412 & 0.330 \\
         Auto COT & 0.523 & 0.311 & 0.365 & 0.346 & 0.282 \\
        
       {\textbf{GPT 4o-mini}} \\
       \fshot{} & 0.625 & 0.451 & 0.501 & 0.595 & 0.533\\ 
       \knn{} \fshot{} & 0.620 & 0.487 & 0.525 & 0.610 & 0.550\\
       \COT{} & 0.645 & 0.459 & 0.513 & 0.652 & 0.578 \\
       Auto COT & 0.524 & 0.384 & 0.420 & 0.546 & 0.481\\ 
       
        \multicolumn{6}{>{\columncolor[gray]{0.93}}l}{\textbf{Open Domain }} \\
        {\textbf{Mistral v0.2 7b}} \\
         BM-25 & 0.746 & 0.547 & 0.607 & 0.623 & 0.581 \\
         all-mini-lm-v2 & 0.572 & 0.380 & 0.430 & 0.414 & 0.366 \\
          mutli-qa-mpnet & 0.552 & 0.380 & 0.441 & 0.436 & 0.375 \\
               {\textbf{GPT-4o-mini}} \\
        BM-25 & 0.735 & 0.666 & 0.685 & 0.751 & 0.729 \\
        all-mini-lm-v2 & 0.621 & 0.528 & 0.551 & 0.627 & 0.578\\
        multi-qa-mpnet & 0.581 & 0.490 & 0.510 & 0.593 & 0.535 \\
        \multicolumn{6}{>{\columncolor[gray]{0.93}}l}{\textbf{Golden Evidence }} \\
        {\textbf{Mistral v0.2 7b}} \\
         \fshot{} & 0.918 & 0.770 & 0.818 & 0.747 & 0.715 \\  
         \knn{} \fshot{} & 0.941 & 0.850 & 0.882 & 0.857 & 0.820\\ 
        {\textbf{GPT-4o-mini}} \\
        \fshot{} &  0.916 & 0.917 & 0.911 & 0.950 & 0.937 \\ 
        \knn{} \fshot{} & 0.937 & 0.950 & 0.934 & \textbf{0.960} & \textbf{0.949} \\ 
        \bottomrule
    \end{tabular}

    \label{tab:main_result}
\end{table*}

\subsection{LLM Evaluation Setups}
We carry LLM evaluation in (1) \textbf{Gold Evidence} setup where, the ground truth infobox is employed to get an upper bound for their performance,  
(2) \textbf{Closed book} setup without external knowledge.
 (3) \textbf{Open-Domain setup} where, we retrieve the relevant evidence (infobox) from the Wikipedia collection (Section \ref{sec:corpus_creation}). We consider two scenarios, one where the documents indexed contain \textit{title, infobox, summaries} (title.infobox-summary (T-I-S) setup) and another where the documents contain only title and summaries (title-summary (T-S) setup). In the T-S setup, a mapping between Wikipedia titles and infoboxes is created once and used for lookup during retrieval, to fetch the corresponding infoboxes for retrieved titles.

\noindent \textbf{Hyperparameters}: For all experiments, we use a temperature of 0.3 decided based on tuning on validation set. 
\vspace{-0.6em}
\subsection{Evaluation metrics}
\textbf{Retrieval}: Mean Reciprocal Rank (MRR)  and Recall@k. 

\noindent \textbf{List Construction Metrics}: To evaluate list construction performance of LLM outputs, we apply the standard definitions of $Precision$, $Recall$, and $F_{1}$ by matching entities in the generated list of answers and the ground truth. To match entities, we follow a three stage pipeline, namely \emph{subsequence} matching and also semantic matching by employing BERT based tokenwise similarity.





\noindent \textbf{Temporal Metrics}: 
Formally, given the set of years covered by the expected date ranges \( Y_{\text{expected}} \) and the set of years covered by the generated date ranges \( Y_{\text{generated}} \), we define Temporal Overlap as: 
$
Temporal\text{ }Overlap = \frac{|Y_{\text{expected}} \cap Y_{\text{generated}}|}{|Y_{\text{expected}}|}
$


The Temporal Jaccard measures the similarity between the generated and expected date ranges. Unlike Temporal Overlap, Temporal Jaccard penalizes the model for \textit{hallucinating} years outside the expected range. Formally, we define it as:
$
Temporal\text{ }Jaccard = \frac{|Y_{\text{expected}} \cap Y_{\text{generated}}|}{|Y_{\text{expected}} \cup Y_{\text{generated}}|}
$
We first parse and normalize the date ranges from both the generated and expected answers. For each matched entity, we represent the date ranges as set of years, which we further use to calculate the Temporal Overlap Score and Temporal Jaccard Similarity. We then average these scores across all matched entities.

\section{Results}
\label{sec:results}
\label{section:standardpromtping}

\subsection{Performance of LLMs on TLQA}
To answer \textbf{RQ 1}, we study the performance of Mistral v0.2 and GPT-4o-mini in a closed book setting on TLQA as shown in Table \ref{tab:main_result} and on \name{}-TS, \name{}-TM in Table \ref{tab:tlqa_hard}. 
From results, we observe that all models have lower listwise recall scores in a closed book setting but have relatively higher precision. We analyzed the answers and observed that this is primarily because the models generate correct answers but incomplete answer lists. We posit that this is primarily due to limitations of model parametric knowledge and  additionally instruction tuned models are more precise but however have lower recall, as evidenced by the study \cite{le2024exploring}. While \cite{le2024exploring} focuses on free-form answers, precision and recall for our work is measured for lists, providing new insights for list construction abilities of LLMs.
Nevertheless, observed behavior is similar: models
fine-tuned to follow human instructions tend to lose completeness in their answers. 

\begin{table*}[h!]
\caption{Performance on TLQA for different Retrieval approaches at k = 1, 3, and 10 on mistral v0.2.}
\small
\centering
\begin{tabular}{llcccccccccc}
\toprule
\multirow{2}{*}{\textbf{Retrieval Setup}} & \multirow{2}{*}{\textbf{Corpus}} & \multicolumn{3}{c}{\textbf{F1@k}} & \multicolumn{3}{c}{\textbf{Temporal overlap@k}} & \multicolumn{3}{c}{\textbf{Temporal Jaccard@k}} \\
\cmidrule{3-11}
 & & \textbf{k=1} & \textbf{k=3} & \textbf{k=10} & \textbf{k=1} & \textbf{k=3} & \textbf{k=10} & \textbf{k=1} & \textbf{k=3} & \textbf{k=10} \\
\midrule
\multirow{2}{*}{all-mini-lm-v2} & T-S & 0.367 & 0.430 & 0.417 & 0.372 & 0.414 & 0.400 & 0.323 & 0.366 & 0.351 \\
 & T-I-S & 0.394 & 0.398 & 0.368 & 0.410 & 0.398 & 0.340 & 0.360 & 0.341 & 0.293 \\
\midrule
\multirow{2}{*}{BM25} & T-S & 0.502 & 0.594 & 0.618 & 0.503 & 0.594 & 0.627 & 0.458 & 0.556 & 0.584 \\
 & T-I-S & 0.536 & 0.607 & \textbf{0.624} & 0.558 & 0.623 & \textbf{0.651} & 0.513 & 0.581 & \textbf{0.607} \\
\midrule
\multirow{2}{*}{multi-qa-mpnet} & T-S & 0.353 & 0.379 & 0.376 & 0.373 & 0.375 & 0.354 & 0.310 & 0.318 & 0.300 \\
 & T-I-S & 0.441 & 0.426 & 0.400 & 0.436 & 0.407 &  0.388 & 0.375 & 0.352 & 0.326 \\
\bottomrule
\end{tabular}

\label{table:rag_performance_at_k}
\end{table*}

\begin{figure}[hbt!]
    \includegraphics[height=0.72\linewidth]{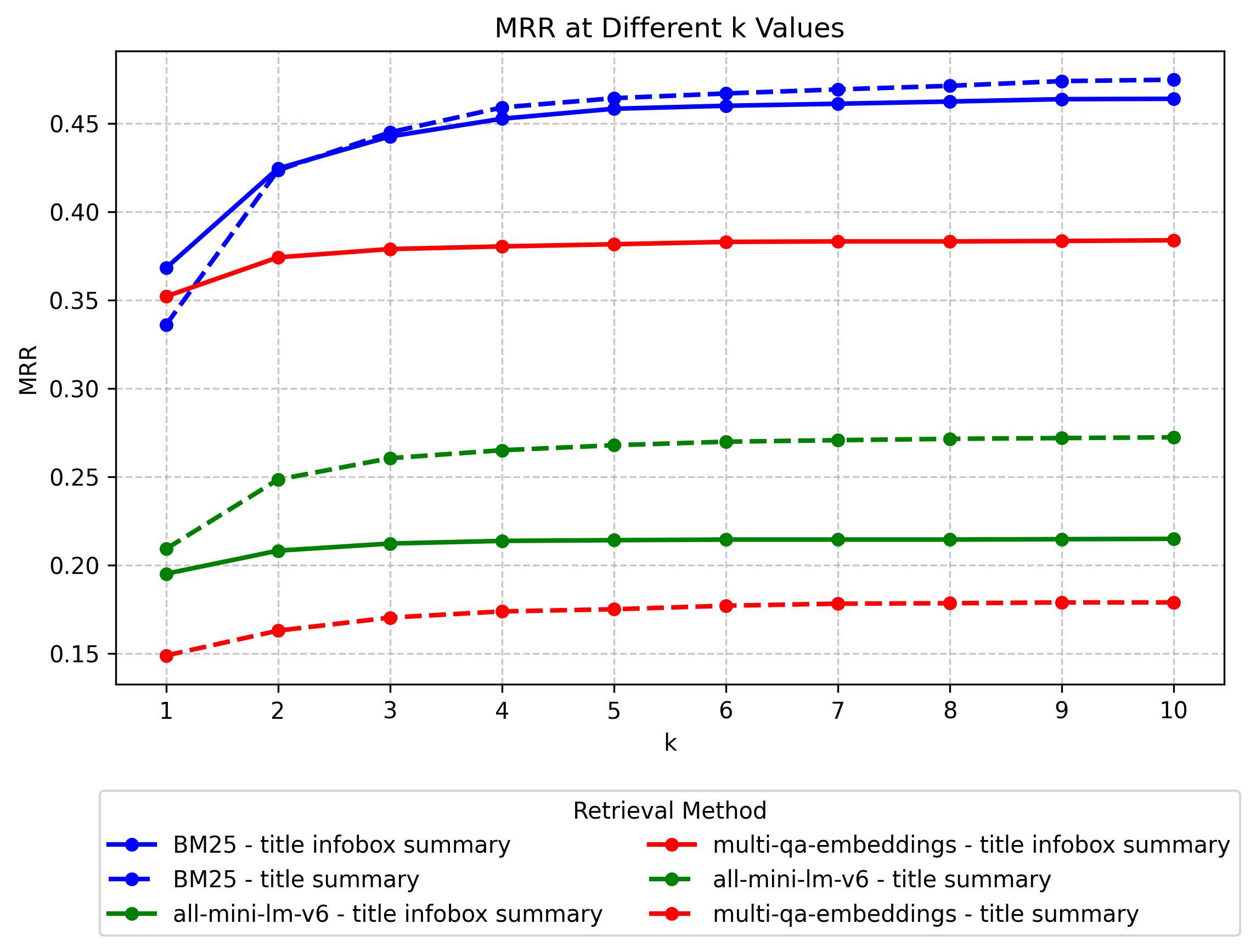}
    \caption{Performance of different retrieval settings over the two Wikipedia corpora, T-S and T-I-S using MRR metric. \textbf{Dashed line} represents T-S, whereas complete line represents T-I-S. }
    \label{fig:MRR_performance}
\end{figure}
\begin{figure}[h]
    \centering
    \includegraphics[width=0.9\linewidth]{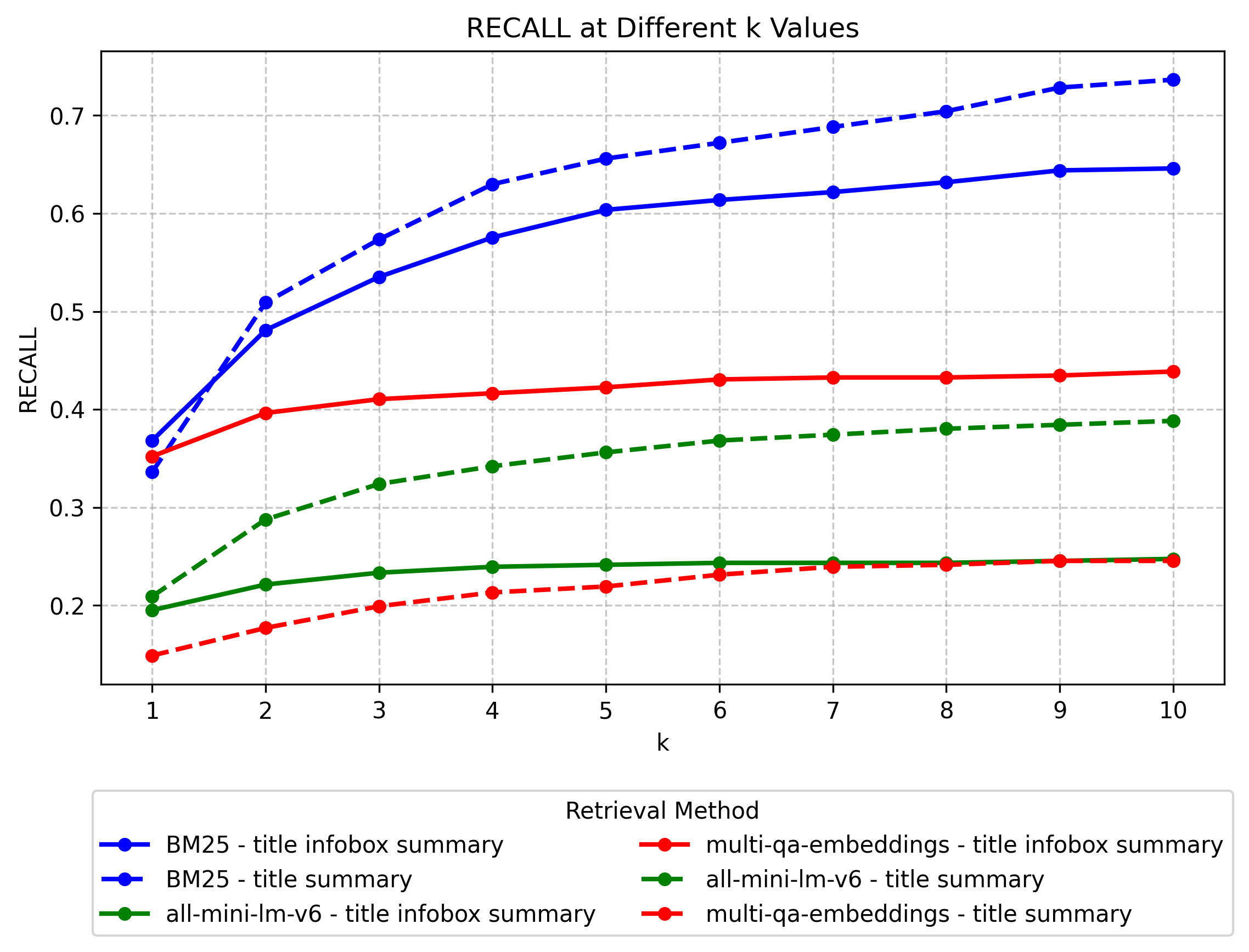}
    \caption{Comparison of retrieval models (Recall) }
                \label{fig:Recall_performance}

\end{figure}

\begin{figure*}[h!]
    \centering
    \includegraphics[width=0.85\linewidth]{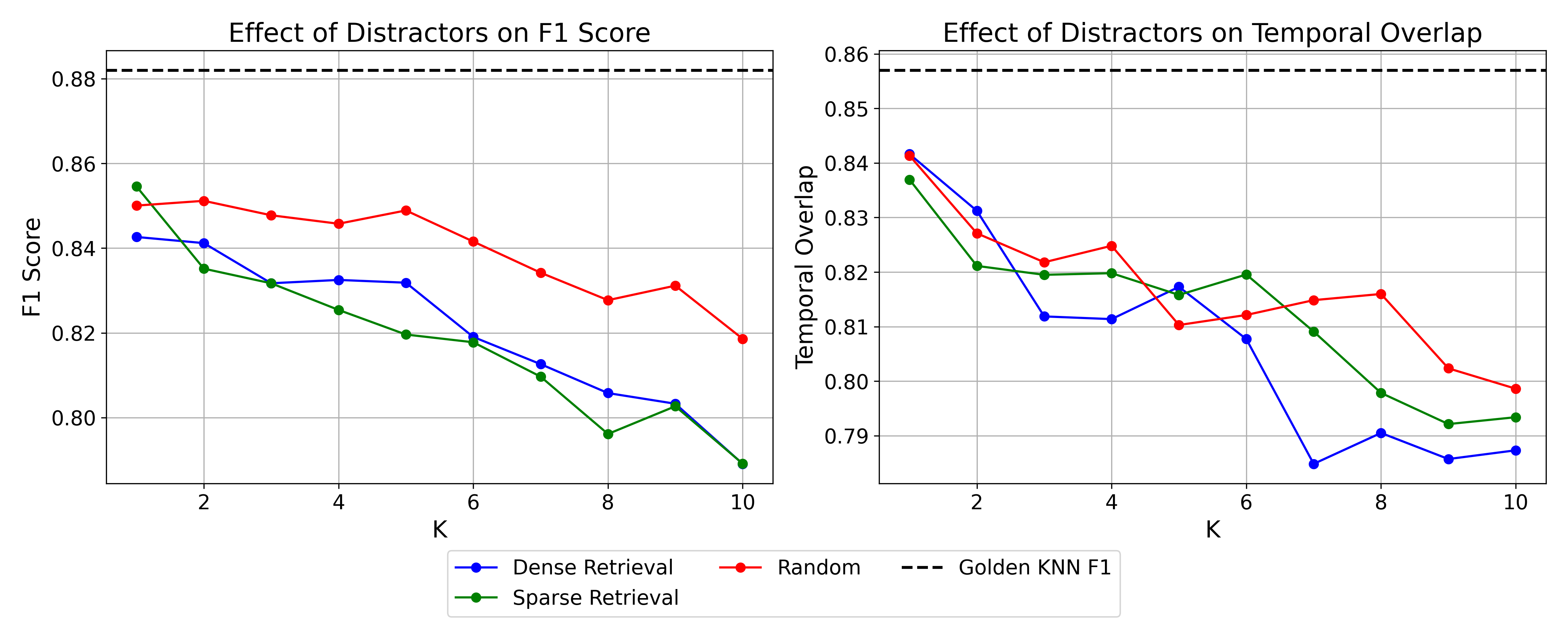}
    \caption{Comparison of F1 Score and Temporal Overlap for answer generation with golden evidence augmented with varying numbers of distractors (k). The dashed line represents the score with golden evidence. Evaluation was conducted on Mistral v0.2 }
    \label{fig:knn_golden_distractors}
\end{figure*}

With respect to \textbf{temporal performance}, both models generally perform suboptimally on TLQA and TM, TS subsets. It can be observed that the Time Overlap scores are consistently higher than the Time Jaccard scores, indicating that while the models are not very capable of identifying significant overlapping time periods, they struggle even more with accurately bounding the time-intervals (start and end periods). This is also evident in the results obtained on \name{}-TS subset that requires temporal bounding abilities. This discrepancy likely arises because the models can detect relevant time frames but have difficulty in precisely delineating the start and end years, resulting in either overshooting (example: Table \ref{tab:incomplete_answer_rob_TM}) or undershooting the correct periods. We also observe that temporal performance on \name{}-TM is the lowest, as the model is unable to detect the implicit temporal reference made by the marker in the question due to parametric memory limitations (example: Table \ref{tab:incomplete_answer_rob_TM}).
Among different prompting methods, \knn{} \fshot{} works best. 

\subsection{Open-Domain Results} 

\textbf{Golden Evidence Setup Results}: To ascertain the upper bound of performance on \name{} and other subsets, we carry out experiments using LLMs in golden evidence setting where the models are provided with the correct evidence. We observe that this leads to higher performance (Table \ref{tab:main_result}, Table \ref{tab:tlqa_hard}) as measured by  list construction metrics (F1=0.934) and temporal metrics (0.960). Hence, better tabular retrieval will lead to high LLM reasoning performance.

\noindent \textbf{Retrieval Performance:}
\label{section:retrieval_performance}
To address \textbf{RQ 2}, we study the impact of augmenting the query with additional evidence to determine if this improves the model's performance on \name{}. We first evaluate the performance of the different retrieval models in open-domain setup on the two Wikipedia corpora (T-S and T-I-S). The performance of the different retrieval systems (MRR) is shown in Figure \ref{fig:MRR_performance} and comparison using Recall@k is shown in Figure \ref{fig:Recall_performance}


From the plot, we observe that sparse retrieval approaches like BM25 outperform dense retrieval approaches on both T-S and T-I-S corpora settings. We posit that this is due to the nature of entity centric queries. Sciavolino et al. \cite{sciavolino2021simpleDenseRetrievalEntity} demonstrate that dense retrieval systems perform considerably worse than BM25 in simple entity-centric questions due to popularity bias.  
\begin{table*}[hbt!]
\small
\centering
\caption{Results for Temporal Markers and Time Intervals Across different setups. model: gpt-4o-mini.}
\begin{tabular}{@{}lcccccc@{}}
\toprule
\textbf{Dataset} & \textbf{Setup} & \textbf{Precision} & \textbf{Recall} & \textbf{F1} & \textbf{TO} & \textbf{TJ} \\ \midrule
\multirow{3}{*}{TLQA-TM} & Closed Book (KNN) & 0.449 & 0.524 & 0.455 & 0.589 & 0.529 \\
 & Open Book (BM25@3) & 0.703 & 0.778 & 0.713 & 0.820 & 0.798 \\
 & Golden Evidence & 0.830 & 0.938 & 0.857 & 0.945 & 0.934 \\ \midrule
\multirow{3}{*}{TLQA-TS} & Closed Book (KNN) & 0.631 & 0.485 & 0.523 & 0.703 & 0.589 \\
 & Open Book (BM25@3) & 0.739 & 0.656 & 0.670 & 0.837 & 0.719 \\
 & Golden Evidence & 0.892 & 0.801 & 0.823 & 0.963 & 0.846 \\ \bottomrule
\end{tabular}

\label{tab:tlqa_hard}
\end{table*}
\begin{table*}[h!]
    \caption{Example of incorrect and incomplete answer from TLQA-TM and TLQA-TS.}
    \centering
    \small
    \renewcommand{\arraystretch}{1.1}
    \begin{tabular}{p{7.5cm} p{6cm}}
    \toprule
   \multicolumn{2}{l}{
        \ding{228} \textbf{Query (TLQA-TM)}: 
        \parbox[t]{15cm}{\ttfamily Which  positions did Rob Nicholson hold after serving as 
        Minister of National Defence?}
    } \\
    \midrule
    \textbf{Expected Answer} & \textbf{Generated Answer} \\
    \midrule
    \begin{itemize}[leftmargin=*,topsep=0pt,parsep=0pt,itemsep=2pt]
        \item Shadow Minister of Justice Shadow Attorney General of Canada: 2015-2019
        \item \textcolor{teal}{Minister of Foreign Affairs: 2015}
    \end{itemize}
    &
    \begin{itemize}[leftmargin=*,topsep=0pt,parsep=0pt,itemsep=2pt]
        \item \textcolor{teal}{Minister of Foreign Affairs: 2015}
        \item \textcolor{red}{Minister of Justice and Attorney General of Canada: 2013-2015}
        \item \textcolor{red}{Minister of National Defence: 2013-2013}
    \end{itemize} \\
    \hline
       \multicolumn{2}{l}{\ding{228} \textbf{Query (TLQA-TS)}: \texttt{Which teams did David Beckham play for between 2000 and 2010?}} \\
    \midrule
    \begin{itemize}[leftmargin=*,topsep=0pt,partopsep=0pt,parsep=0pt,itemsep=0pt]
        \item Manchester United F.C.: 2000-2003
        \item \textcolor{teal}{Real Madrid CF: 2003-2007}
        \item \textcolor{teal}{England national football team: 2000-2009}
        \item \textcolor{teal}{LA Galaxy: 2007-2010}
        \item \textcolor{teal}{AC Milan (loan): 2009, 2010}
    \end{itemize}
    
    &
    \begin{itemize}[leftmargin=*,topsep=0pt,partopsep=0pt,parsep=0pt,itemsep=0pt]
        \item \textcolor{teal}{Real Madrid: 2003-2007}
        \item \textcolor{teal}{LA Galaxy: 2007-\textcolor{red}{2012}}
        \item \textcolor{teal}{England national football team: 2000-2009}
        \item \textcolor{teal}{AC Milan (loan): 2009-2010}
    \end{itemize} \\
        \midrule

    \end{tabular}

    \label{tab:incomplete_answer_rob_TM}
\end{table*}

\noindent \textbf{Retrieval Augmented Answer Generation}: We also observe the impact of retrieved documents in an open-domain setting on downstream LLM reasoning, as shown in Table \ref{tab:main_result} for \name{}. We observe that augmenting LLMs with retrieved documents in general leads to performance gains in list construction and temporal metrics. We observe that retrieval performance directly translates to answer generation performance \cite{leto2024optimalsearchretrievalrag}, as BM25 retrieved documents lead to significantly higher gains compared to documents from dense retrieval. For instance, BM25 documents coupled with Mistral v0.2 outperforms all-mini-lm retrieved documents by \textbf{45.11\%} as measured by $F_1$. The results in Table \ref{tab:main_result} for open-domain setting are carried out with top-3 retrieved documents. We further vary k=1,3,10 (number of retrieved documents) and study their impact on downstream LLM based answer generation as shown in Table \ref{table:rag_performance_at_k}. We observe that LLM reasoning using documents from sparse and dense retrieval approaches, results in slight gains and saturated performance respectively beyond k=3. Further, we observe that LLMs are able to decipher implicit temporal references in \name{}-TM when provided with the relevant infobox by mapping the marker to time interval information. However, analysis on \name{}-TS reveals that the temporal bound error persists. 

\begin{table*}[h!]
    \caption{Example of refusal errors and incomplete answers. }
    \small
    \begin{tabular}{p{7.4cm} p{7.4cm}}
    \toprule
    \multicolumn{2}{l}{\ding{228} \textbf{Query}: \texttt{List all teams  Alan Kirby played to this day.} \textbf{Refusal Error}} \\
    \midrule
    \textbf{Expected Answer} & \textbf{\textbf{Generated Answer} } \\
    \midrule

    \begin{itemize}[leftmargin=*,topsep=0pt,partopsep=0pt,parsep=0pt,itemsep=0pt]
        \item Aston Villa F.C.: 1996-1998, 1994-1996
        \item Johnville F.C. (Waterford): 1994
        \item Republic of Ireland under-21 national football team: 1998
    \dots
    \end{itemize}

            & I apologize, but I couldn't find any information on a person named Alan Kirby playing for any football teams. If you meant someone else or provided incorrect information, please let me know and I'll be happy to help.  \\
    
    \bottomrule
    \multicolumn{2}{l}{\ding{228}
    \textbf{Query}: \texttt{List all political positions Ulf Kristersson held to this day.} \textbf{Incomplete Answer Error}}  \\
    \midrule
    \textbf{Expected Answer} & \textbf{Generated Answer} \\
    \midrule
    \begin{itemize}[leftmargin=*,topsep=0pt,partopsep=0pt,parsep=0pt,itemsep=0pt]
        \item Prime Minister of Sweden: 2022-2024
        \item \textcolor{teal}{Leader of the Moderate Party: 2017-2024}
        \item Leader of the Opposition: 2017-2022
        \item Minister for Social Security: 2010-2014
        \item Chairman of the Moderate Youth League: 1988-1992
        \item \textcolor{teal}{Member of the Riksdag: 2014-2024, 1991-2000}
    \end{itemize}
    
    &
    \begin{itemize}[leftmargin=*,topsep=0pt,partopsep=0pt,parsep=0pt,itemsep=0pt]
    \item \textcolor{teal}{Leader of the Moderate Party: 2017-2024}
    \item \textcolor{teal}{Member of the Riksdag:}\textcolor{red}{ 2006-2021, 2022-2024}
    \end{itemize}
    
    \\
    \bottomrule
    \end{tabular}

    \label{tab:refusal_example}
    
\end{table*}

\vspace{-1em}
\subsection{Effects of Distractors}
To address \textbf{RQ 3}, we evaluate how different types of distractors impact the performance of the golden evidence based LLM reasoning setup. This would also provide insights into future directions for improving retrieval. We use three different methods for selecting distractors: random selection, retrieving the top $k$ infoboxes using BM25, or using dense retrieval with all-mini-lm-v2 embeddings with T-S corpora setup.  We then append the golden evidence along with these $k$ distractors as evidence for the LLM.

From Figure \ref{fig:knn_golden_distractors}, it can be observed that temporal and list construction performance drops noticeably with the addition of even a single distractor ($k=1$), indicating that noise affects the model's ability to extract the correct information to answer the query. This effect becomes more evident as we increase the number of distractors ($k$). This consistent performance degradation can be attributed to the model's capacity limitations; as more distractors are introduced, it becomes difficult for the model to discern useful information from the correct evidence. Hence, we observe that the model outputs incorrect entities or is unable to output correct time periods due to presence of distracting information. Further, we observe that randomly choosing distractors leads to a less significant decrease in performance compared to distractors retrieved using BM25 or dense retrieval model. We posit that this maybe because the retrieved distractors serve as hard negatives compared to random distractors, rendering it difficult for the LLM to distinguish between golden evidence and the distractors. These results contrast the phenomenon observed recently \cite{power_of_noise}, where the authors observe that adding random noise to contexts improves performance of RAG systems.

\vspace{-1em}
\subsection{Error Analysis}
\label{sec:error_analysis}

We perform a detailed error analysis of common errors made by LLMs in our evaluation on the test sets. From Table \ref{tab:refusal_example}, we observe the case of \textit{refusal errors} where the model is not able to generate an answer in the question due to limitations of parametric knowledge. LLMs are believed to encode world knowledge in their parameters and are believed to perform \textit{approximate retrieval} \cite{Kambhampati_2024} when posed with new queries. However, we observe that LLMs are unable to provide an answer in some cases as discussed above, either due to limitations in parametric knowledge or inability of the LLM to perform approximate retrieval in closed book setting. We observe that in a retrieval augmented setting for LLMs the \textbf{refusal errors} decrease when related evidence is present among retrieved infoboxes.


In Table \ref{tab:refusal_example}, we also observe  \textit{incomplete answer} errors and \textit{incorrect temporal bounds}. We observe that the model only covers two positions, ``leader of moderate party" and ``member of Riksdag" out of 6 positions held by the individual in question. We posit that incomplete answers could again be an artifact of limited knowledge encoded in model parameters in a closed book setting. We also observe that in many scenarios where the LLM generates incomplete answers, it is also an artifact of \textit{popularity bias} as the LLM covers popular organizations that are well known and less popular or rare entities/organizational names are left out.

We also observe from Table \ref{tab:refusal_example} that  the answer ``Member of Riksdag" has wrong temporal bounds. We observe that LLM generated time interval has \textit{undershooting errors} where it predicts time period as starting from 2006 when in actuality it was 2014 and similar error for the end time period (2021 vs 2024). it also has an \textit{overshooting problem} where the model predicts 2022 instead of 1991. We observe the \textit{overshooting} and \textit{undershooting} issues for time intervals in closed book and even in open-domain settings,  demonstrating that LLMs still lack at accurate temporal understanding.

We also perform error analysis on evaluation subsets \name{}-TS and \name{}-TM as shown in Table \ref{tab:incomplete_answer_rob_TM}. We observe from Table \ref{tab:incomplete_answer_rob_TM} that the LLM is unable to detect and reason about the implicit temporal reference made by  the temporal marker ``after serving  as Minister of National Defense" in the question. This results in the LLM  generating positions held by ``Rob Nicholson" prior to the time period mentioned in the question. Due to incorrect temporal understanding, it also results in missing a position held by this person. We also observe a case of \textbf{incorrect temporal bounds}, where the LLM is not able to detect the bounds of the time span mentioned in the question and overshoots the end time period for the ``La Galaxy".
\vspace{-1.7em}
\section{Conclusion}
We introduce \name{}, an open-domain benchmark where the models are expected to provide a list of possible answers with corresponding time intervals. \name{} primarily tests the temporal understanding and list construction abilities of models. We observe that LLMs believed to encode world knowledge underperform by either predicting incomplete lists or suffer from temporal understanding blindspots resulting in incorrect time intervals or temporal alignment. While we observe that RAG setup improves performance, there exists scope for improvement of tabular retrieval and temporal understanding capabilities of LLMs, providing future research directions.
\section{Acknowledgments}
We would like to thank Sowmya AS for helping with qualitative analysis of questions generated in \name{} (Section 3.2, Table \ref{tab:question_quality}). We also would like to thank the reviewers for their valuable feedback.

\appendix

\bibliographystyle{ACM-Reference-Format}
\balance
\bibliography{bib}


\end{document}